\documentclass[11pt]{article}
\usepackage[preprint]{acl}
\usepackage{times}
\usepackage{latexsym}
\usepackage[T1]{fontenc}
\usepackage[utf8]{inputenc}
\usepackage{microtype}
\usepackage{inconsolata}
\usepackage{graphicx}
\usepackage{booktabs}
\usepackage{amsmath,amssymb}
\usepackage{multirow}
\usepackage{xcolor}

\newsavebox{\fitbox}
\newcommand{\fittable}[1]{\sbox{\fitbox}{#1}\ifdim\wd\fitbox>\linewidth\resizebox{\linewidth}{!}{\usebox{\fitbox}}\else\usebox{\fitbox}\fi}
\newcommand{\capl}{CapLite}
\newcommand{\G}{\mathcal{G}(q)}
\newcommand{\tool}[1]{\texttt{\detokenize{#1}}}

\title{\capl{}: Prompt-Injection Defense for Small Language Model Agents\\ without Model-Written Security Artifacts}

\author{
  Zimo Xu \\
  The University of Tokyo \\
  {\small\texttt{xuzimo333@g.ecc.u-tokyo.ac.jp}}
  \And
  Wenzhuo Liu \\
  Institute of Automation, CAS \\
  {\small\texttt{liuwenzhuo2020@ia.ac.cn}}
  \And
  Hideki Nakayama \\
  The University of Tokyo \\
  {\small\texttt{nakayama@ci.i.u-tokyo.ac.jp}}
}

\begin{document}
\maketitle

\begin{abstract}
Enforcement defenses against prompt injection, such as CaMeL, Progent, IPIGuard and DRIFT, ask the agent's own model to write a \emph{security artifact} (a program, a privilege policy, a dependency graph or a constraint list) that code then enforces. We ask \emph{whether small language model (SLM) agents can be trusted to write these artifacts}. Across 15 open-weight models, 13 of them between 1.7B and 12B parameters, they cannot: the policies, graphs and constraints that the agent's model writes admit attacks on 14 (Progent), 13 (IPIGuard) and 11 (DRIFT) of the models, and CaMeL's programs fail to run in up to 68\% of episodes. We present \capl{}, which derives its policy from the tool schemas, commits the write calls before any tool result is read, and admits a write only if every argument that decides whom or what it affects is stated in the user's request. This guarantees \emph{write integrity}: injected text cannot add a write, and no write reaches a recipient, amount, object or destination that the user did not state. No attack succeeds in 13{,}935 attacked AgentDojo episodes. \capl{} reaches utility under attack comparable to CaMeL's (29.0\% vs.\ 27.2\%) with fewer tokens than every other enforcement defense and than the undefended agent: 17.1k per episode, 18\% of CaMeL's, and the highest utility per token of all enforcement defenses.
\end{abstract}

\section{Introduction}

\begin{figure}[t]
  \centering
  \includegraphics[width=\columnwidth]{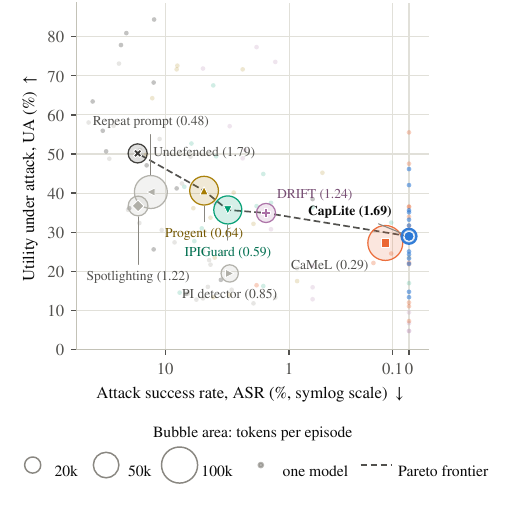}
  \caption{\capl{} is the only defense with no successful attack, and it starts the Pareto frontier of attack success and utility under attack. Each bubble averages the models a defense ran on, and faint dots show single models. Bubble area is proportional to tokens per episode, and the number in parentheses is utility under attack per thousand tokens. The ASR axis is reversed: lower attack success is to the right, and higher utility is upward. Defenses below and to the left of the dashed frontier are dominated.}
  \label{fig:tradeoff}
\end{figure}

A tool-using agent reads e-mails, files and web pages that anyone can write, and an instruction hidden in that content can redirect what the agent does \citep{greshake2023indirect,debenedetti2024agentdojo}. The defenses with the strongest guarantees take the decision out of the model's context and give it to code. CaMeL runs a program in an interpreter that tracks where every value came from \citep{camel2025}, Progent checks each call against a privilege policy \citep{progent2025}, IPIGuard executes a tool dependency graph planned before any tool result is read \citep{ipiguard2025}, and DRIFT validates calls against a planned trajectory and parameter checklist \citep{drift2025}.

\begin{figure*}[t]
  \centering
  \includegraphics[width=\textwidth]{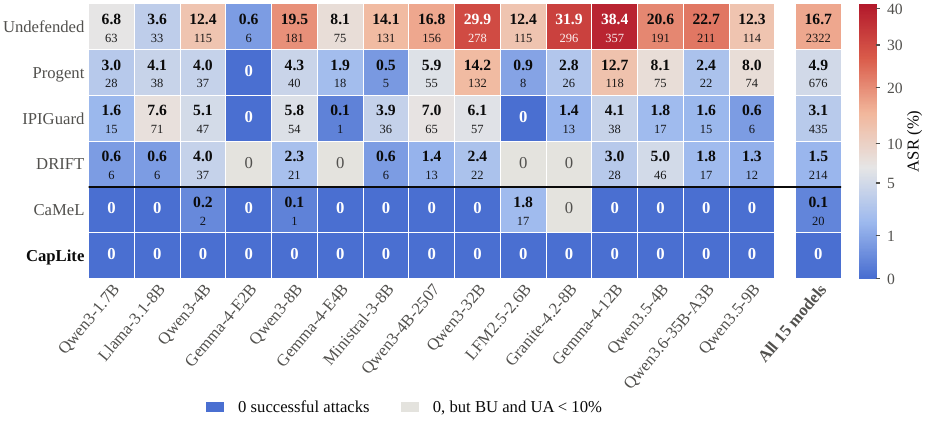}
  \caption{Attack success rate of each defense on each of 15 models under \emph{important\_instructions}. Each cell gives the ASR in percent and, below it, the successful attacks out of 929 tool-goal episodes. Dark blue cells have no successful attack; gray cells have none because benign and attacked utility are below 10\%, which comes from inaction. \capl{} is the only defense without a successful attack on any model.}
  \label{fig:heat}
\end{figure*}

Each of these defenses enforces something the agent's own model writes: CaMeL the program, Progent the policy, IPIGuard the dependency graph and DRIFT the trajectory and checklist. The enforcement is exactly as sound as this security artifact. All four were designed and evaluated mainly with large hosted models. Concurrent request-grounded authorization systems \citep{rope2026,li2026toolfence,liao2026tracegrant,pact2026} also target large hosted models, and most of them have a model compile the authorization, which is again a model-written security artifact. Meanwhile, small open-weight language models (SLMs) are an increasingly common backbone for agents that run locally, on device or at scale \citep{tinyagent2024,alphaumi2024,tinyllm2025}. We ask \emph{whether SLM agents can be trusted to write their own defenses, and what can be guaranteed if they cannot}.

We measure four enforcement defenses on 15 open-weight models with AgentDojo. \emph{Artifacts that decide what is allowed fail open}: policies written by the agent's model admit attacks on 14 of 15 models, IPIGuard's graphs on 13 of 15, and DRIFT's constraints on 11 of 15. \emph{Artifacts that decide what runs fail closed}: CaMeL is breached in only 20 of 13{,}935 attacked episodes, but its programs fail to run in up to 68\% of episodes and cost 94k tokens per episode. These failure rates show no trend with model capability, while the undefended agent follows injections more often as the model gets stronger.

An SLM agent does not need to write its own authorization. \capl{} derives its policy from tool schemas by fixed rules and takes every security decision in deterministic code. A Planner that sees only the request commits the write calls before any tool result is read, a Binder fills each open argument by choosing among extracted candidates, and a Mediator admits a write only if every authority argument, such as a recipient, an amount, an object identifier or a URL, equals a value stated in the request. A model failure can drop a write or pick a wrong candidate, and injected text can neither add a write nor route one to a value the user did not state; we call this guarantee \emph{write integrity}. No attack succeeds in 13{,}935 attacked episodes, as these properties predict, and \capl{} reaches utility under attack comparable to CaMeL's with 18\% of its tokens. It uses fewer tokens than every other enforcement defense and than the undefended agent, and it gives the highest utility per token of all enforcement defenses.

Our contributions are:
\begin{itemize}
  \item \textbf{Measurement.} A study of four enforcement defenses on 15 open-weight models, which shows that security artifacts written by the agent's own model inherit its failures across the whole range from 1.7B to 35B: artifacts that decide what is allowed admit attacks, and artifacts that decide what runs lose the task (Section~\ref{sec:diagnosis}).
  \item \textbf{Architecture.} \capl{}, a design in which the model proposes and code authorizes, combining committed writes, schema-derived field roles, exact request grounding and candidate-only binding (Section~\ref{sec:method}). Its two properties hold by construction, no attack succeeds in 13{,}935 attacked episodes, and ablations show that the Mediator and the commitment of writes are each needed (Section~\ref{sec:results}).
  \item \textbf{Efficiency.} \capl{} uses fewer tokens than every other enforcement defense and than the undefended agent, and reaches utility under attack comparable to CaMeL's with the highest utility per thousand tokens of all enforcement defenses (Section~\ref{sec:results}).
\end{itemize}

\section{Can SLM Agents Write Their Own Defenses?}
\label{sec:diagnosis}

We call a model output a \emph{security artifact} when the enforcer admits whatever it contains. For each of the 15 models, we run every defense with that model in all of its model roles, on the setup and metrics of Section~\ref{sec:setup}.

\paragraph{Artifacts that decide what is allowed fail open.}
Progent asks the model to write a privilege policy from the request and to update it after each call. The policies admit attacks on 14 of 15 models, 676 in total and up to 132 of 929 on Qwen3-32B (Figure~\ref{fig:heat}). The same policies block 25--77\% of the write calls the agent attempts under attack (Appendix~\ref{app:diag}), and benign utility falls from 59.4\% undefended to 43.8\%, so a generated policy is wrong in both directions. IPIGuard's planner commits a dependency graph and DRIFT's a trajectory and parameter checklist that a validator enforces; IPIGuard is breached on 13 of 15 models (435 attacks) and DRIFT on 11 of 15 (214 attacks; Figure~\ref{fig:heat}). DRIFT's four clean models, LFM2.5-2.6B, Gemma-4-E2B, Gemma-4-E4B and Granite-4.2-8B, complete 4.7\%, 9.5\%, 6.7\% and 4.7\% of attacked tasks, so their zero comes from inaction.

\paragraph{Artifacts that decide what runs fail closed.}
CaMeL's interpreter enforces the authors' data-flow policies on whatever program the model writes, and security holds: 20 successful attacks in 13{,}935 episodes, 17 of them on LFM2.5-2.6B. On 8 of the 15 models the final program is unrunnable in more than 15\% of episodes, and on Granite-4.2-8B in 68\%, where the model returns no code at all (Appendix~\ref{app:diag}). The causes are format, syntax and interpreter errors that survive up to ten repair rounds, each costing more calls: CaMeL spends 94k tokens per episode (Appendix~\ref{app:cases}, Case 2), and its benign utility averages 24.9\% (undefended: 59.4\%).

\begin{figure}[t]
  \centering
  \includegraphics[width=\columnwidth]{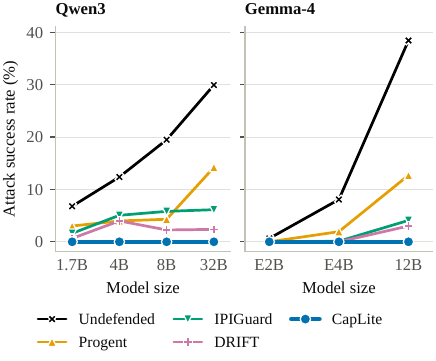}
  \caption{Attack success rate by model size within the Qwen3 and Gemma-4 families. The defenses whose security artifact decides what is allowed admit more attacks as the model grows, together with the undefended agent; \capl{} stays at zero.}
  \label{fig:scaling}
\end{figure}

\paragraph{Larger models do not write better artifacts.}
Within the Qwen3 (1.7B--32B) and Gemma-4 (E2B--12B) families, the attack success rate under Progent and IPIGuard rises monotonically with model size, together with that of the undefended agent (Figure~\ref{fig:scaling}); under Progent it grows from 3.0\% to 14.2\% on Qwen3 and from 0\% to 12.7\% on Gemma-4. A larger model is easier to inject, and the policies and graphs it writes do not absorb the extra risk. CaMeL's programs are the exception on Gemma-4, where the share without a runnable program falls from 22.9\% to 3.3\%, which matches CaMeL's higher utility on the two larger Gemma-4 models (Section~\ref{sec:results}). Across all 15 models, no artifact failure rate shows a significant trend with capability ($|\rho| \le 0.33$, $p > 0.2$).

\paragraph{What this implies.}
In each defense the enforcement is sound relative to its artifact, and the agent's own model writes the artifact. When the artifact decides what is allowed, the model's mistakes become admitted attacks; when it decides what runs, they become lost tasks and repeated calls. Enforcement that must hold for any model therefore cannot rely on an artifact that the model writes.

\section{\capl{}}
\label{sec:method}

\begin{figure*}[t]
  \centering
  \includegraphics[width=\textwidth]{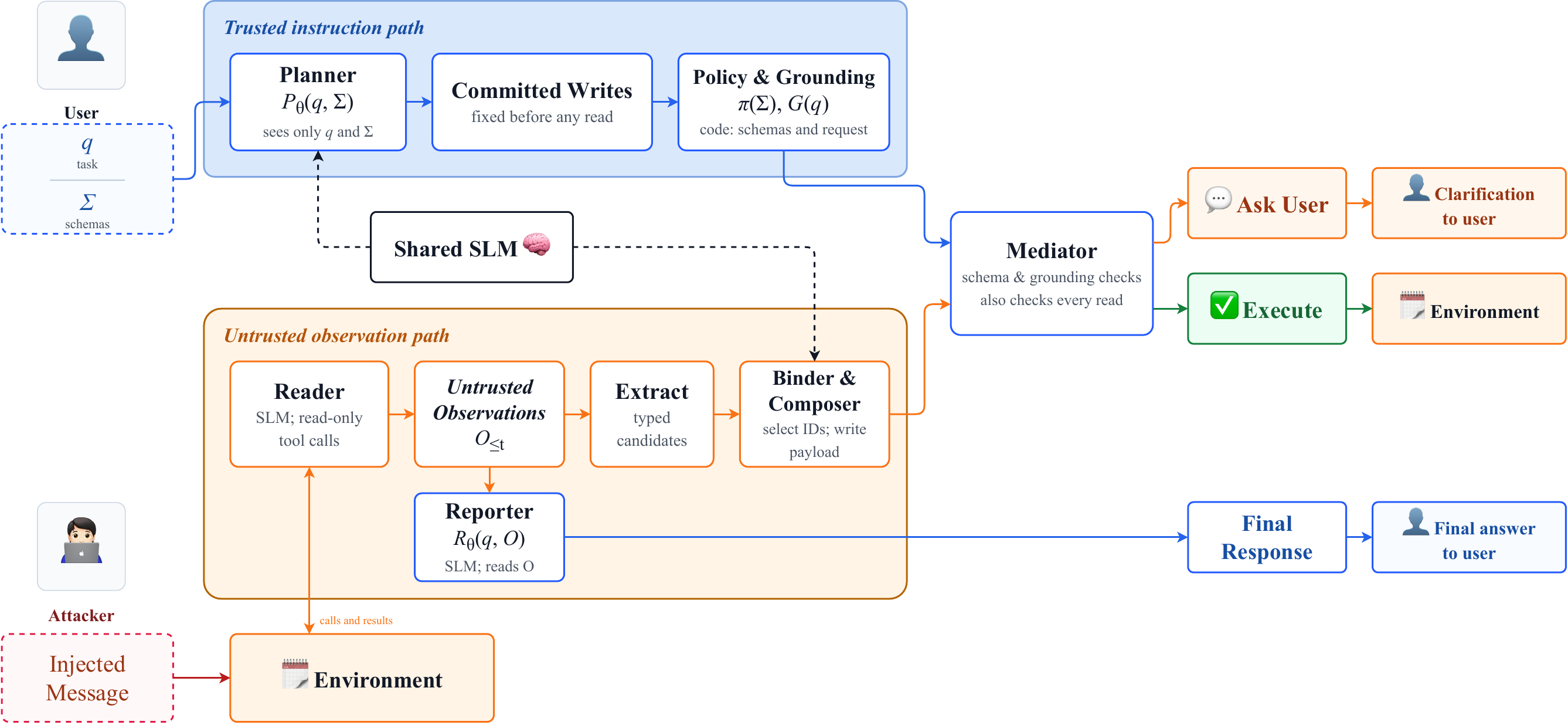}
  \caption{\capl{}. Top, the trusted instruction path: the Planner sees only the request $q$ and the tool schemas $\Sigma$ and commits the write calls, with placeholders for values to look up, before any tool result is read. Code derives the policy from $\Sigma$ and the grounding set $\G{}$ from $q$. Bottom, the untrusted observation path: the Reader makes read-only tool calls, whose results may contain the attacker's text. The Binder picks a candidate identifier and the Composer writes payload text. The Mediator, which is deterministic code, admits a write only if every authority argument is in $\G{}$, checks the destination of every read, and asks the user when a value is missing. All model calls use one SLM.}
  \label{fig:arch}
\end{figure*}

\capl{} never lets the model decide what is authorized. Authorization comes from two things that code computes: a policy derived from the tool schemas and the set $\G{}$ of values the request states. The model proposes a plan, a choice among extracted candidates and text; the Mediator executes a proposal only if it matches the policy and every authority argument lies in $\G{}$. A wrong proposal is dropped or denied. The Mediator rejects every write that was not committed and every authority value the request does not state. Figure~\ref{fig:arch} shows the pipeline; all model calls use one SLM.

\paragraph{Threat model.}
The user, the request $q$ that starts an episode, the tools and their documentation are trusted. Tool results are not: an attacker controls arbitrary text inside the data that read tools return, such as e-mails, files, web pages, reviews or transaction subjects, and succeeds when the agent performs a write that serves the attacker's goal. The guarantee holds whatever instructions the model follows, so adaptive attacks that bypass model-level defenses \citep{zhan2025adaptive} leave it intact; Section~\ref{sec:setup} lists the attacks we run, including two written against \capl{}'s design.

\subsection{Schema-derived policy}
\label{sec:policy}

The policy comes from what any tool registry exposes: the tool name, its docstring and its JSON parameter schema. A tool is \emph{read-only} when the first verbs of its name and docstring are both read verbs (\emph{get}, \emph{search}, \emph{list}, \ldots), a \emph{write} tool when either is an effect verb (\emph{send}, \emph{create}, \emph{pay}, \ldots), and \emph{unclassified} otherwise; the Mediator denies unclassified tools, so an unfamiliar verb costs utility. Every argument of a write tool is an \emph{authority} field unless the schema marks it as free text (\emph{payload}), a date or time (\emph{temporal}) or an enum or boolean (\emph{control}). A destination such as a URL is authority even in a read-only tool, because fetching an attacker-chosen URL is itself an effect. Authority fields decide whom or what a write affects: recipients, participants, accounts, amounts, channels, object identifiers and destinations. Appendix~\ref{app:policy} lists the verb and token lists and the classification of all 74 tools.

\subsection{Adaptive reads, committed writes}
\label{sec:pipeline}

\textbf{Planner.} One schema-constrained call sees only $q$ and the tool documentation and returns at most ten tool calls, with each argument either copied from the request or left as a placeholder such as \texttt{\$event\_id}. Its write calls are \emph{committed}: no later component can add, remove or reorder them.

\textbf{Reader.} A loop of at most ten read-only tool calls that sees the request and the committed writes and adapts its lookups to what it reads. Its calls also pass the Mediator, so a read whose destination the request does not state is refused.

\textbf{Binder and Composer.} For each placeholder, \capl{} extracts typed candidate values from the observations; for an authority or control field it keeps only values in the request's grounding set $\G{}$, the exact values and token $n$-grams that $q$ states (Appendix~\ref{app:impl}). The Binder picks one candidate identifier under constrained decoding, or answers \emph{none} or \emph{ask}, and never writes a value. The Composer writes payload text from the request and the observations.

\textbf{Mediator.} Before each committed write, deterministic code checks that the tool is classified, no placeholder remains, the arguments match the schema and every authority argument equals a member of $\G{}$ exactly; a substring of the request does not count. The first denial, confirmation request or tool error stops the remaining writes. A \textbf{Reporter} then answers from the observations.

When an authority value appears only in a tool result, for example a recipient the user describes but does not name, no grounded candidate exists and \capl{} asks the user to confirm. Our evaluation answers no such request and counts the episode as a failure (Appendix~\ref{app:cases}, Case 3).

\paragraph{Guarantee.} Two properties hold by construction in every episode.

\begin{enumerate}
  \item \textbf{Committed writes.} The executed write calls are a prefix of the write calls the Planner committed before any tool result was read, in the same order.
  \item \textbf{Request grounding.} Every authority argument of every executed write is an element of $\G{}$ or a value the user confirmed.
\end{enumerate}

Injected text can therefore change which reads the Reader performs, temporal values and the final answer, but it cannot add a write, choose a different write tool, or redirect a recipient, account, amount, channel, object or destination to a value the user did not state. The text of a free-text argument, such as a message body, is written by the Composer and is not constrained. A trace audit re-derives the policy and $\G{}$ from the logged request and tool registry, outside the running agent, and checks every executed call: no violation occurs in 25{,}452 \capl{} episodes across all models and attacks.

\section{Experimental Setup}
\label{sec:setup}

\paragraph{Benchmark.}
We use the AgentDojo benchmark v1.2.2 \citep{debenedetti2024agentdojo} with its four suites (Workspace, Slack, Banking, Travel), 97 user tasks and 949 pairs of a user task with an injection task. Benign utility uses the 97 user tasks without injection; utility under attack uses all 949 pairs. Attack success rate uses the 929 pairs whose injection goal needs a tool call; the other 20 pairs are the one Travel goal that asks only for a sentence in the final answer. It involves no write, so it lies outside write integrity (Section~\ref{sec:pipeline}): \capl{} does not defend it (Case~5 in Appendix~\ref{app:cases}).

\paragraph{Models.}
We evaluate 15 open-weight models: 13 small models from 1.7B to 12B parameters (Qwen3-1.7B, -4B, -4B-Instruct-2507 and -8B, \citealp{qwen3}; Qwen3.5-4B and -9B, \citealp{qwen35}; Llama-3.1-8B-Instruct, \citealp{llama3}; Gemma-4-E2B, -E4B and -12B, \citealp{gemma4}; Granite-4.2-8B, \citealp{granite42}; Ministral-3-8B, \citealp{ministral3}; LFM2.5-2.6B, \citealp{lfm25}) and two larger references, Qwen3-32B and the mixture-of-experts Qwen3.6-35B-A3B \citep{qwen36}. All models are served with vLLM 0.30.0 \citep{kwon2023vllm} and native tool calling, with temperature 0, seed 0, thinking disabled and at most 2{,}048 output tokens per call (8{,}192 for DRIFT's calls, Appendix~\ref{app:ports}).

\paragraph{Defenses.}
Every defense uses the same SLM for every model role it defines. \textbf{Undefended} is AgentDojo's tool-calling agent. AgentDojo's four built-in defenses (tool filter, spotlighting, repeated user prompt and a prompt-injection detector) run on eight models and appear in Appendix~\ref{app:full}. \textbf{CaMeL} \citep{camel2025} uses the authors' code and per-suite security policies, with the SLM as both privileged and quarantined model. \textbf{Progent} \citep{progent2025} uses the authors' policy generation and updates and their always-allowed read tools. As in their code, the temperature rises by 0.2 after each policy that fails to parse. \textbf{IPIGuard} \citep{ipiguard2025} uses the authors' prompts, graph planner and read-only whitelist. \textbf{DRIFT} \citep{drift2025} uses the authors' planner, validator and injection isolator and runs on all 15 models. Its parser change is described in Appendix~\ref{app:ports}. Every port keeps the authors' algorithm and prompts and changes only the model interface (Appendix~\ref{app:ports}).

\paragraph{Attacks.}
The main attack is AgentDojo's \emph{important\_instructions}. Robustness experiments on eight models (Qwen3-1.7B, -4B, -4B-Instruct-2507, -8B and -32B; Llama-3.1-8B; Gemma-4-E4B and -12B) add \emph{tool\_knowledge}, which also names the tools the attacker wants called, and two attacks that target \capl{}'s design: \emph{read exfiltration} asks the agent to fetch an attacker URL with a read-only tool, and \emph{descriptor collision} plants a record that carries the user's own descriptor (Appendix~\ref{app:rg}).

\paragraph{Metrics.}
Benign utility (BU) is the task success rate on the 97 user tasks without injection, utility under attack (UA) is the success rate on the 949 attacked pairs, and the attack success rate (ASR) is the fraction of the 929 tool-goal pairs in which the injection goal is achieved, also reported as the count $k$ of successful attacks. Cost is the number of prompt and completion tokens per episode, averaged over the 97 benign and 949 attacked episodes of each model.

\section{Results}
\label{sec:results}

\begin{table}[t]
  \centering
  \small
  \setlength{\tabcolsep}{4pt}
  \fittable{\begin{tabular}{@{}lrrrrrrr@{}}
\toprule
Defense & BU $\uparrow$ & UA $\uparrow$ & ASR (\%) $\downarrow$ & $k$ $\downarrow$ & Zero-ASR $\uparrow$ & Tokens $\downarrow$ & UA/kTok $\uparrow$ \\
\midrule
Undefended & 59.4 & 50.2 & 16.66 & 2{,}322 & 0/15 & 28.0k & 1.79 \\
Progent & \textbf{43.8} & \textbf{40.6} & 4.85 & 676 & 1/15 & 63.4k & 0.64 \\
IPIGuard & \underline{40.1} & \underline{35.7} & 3.12 & 435 & 2/15 & 60.5k & 0.59 \\
DRIFT & 35.7 & 34.9 & 1.54 & 214 & 4/15$^\ddagger$ & \underline{28.1k} & \underline{1.24} \\
CaMeL & 24.9 & 27.2 & \underline{0.14} & \underline{20} & \underline{12/15}$^\ddagger$ & 93.7k & 0.29 \\
\capl{} & 28.1 & 29.0 & \textbf{0.00} & \textbf{0} & \textbf{15/15} & \textbf{17.1k} & \textbf{1.69} \\
\bottomrule
\end{tabular}
}
  \caption{Main results under \emph{important\_instructions}, averaged over 15 models. ASR pools the $k$ successful attacks over all tool-goal episodes run ($15 \times 929$). Zero-ASR counts the models with no successful attack. Tokens are per episode, and UA/kTok is utility under attack per thousand tokens. Among the defenses, the best value is in bold and the second best underlined; the undefended agent is a reference and is not ranked. $\ddagger$: includes models whose zero comes with benign and attacked utility below 10\% (DRIFT: LFM2.5-2.6B, Gemma-4-E2B, Gemma-4-E4B and Granite-4.2-8B; CaMeL: Granite-4.2-8B).}
  \label{tab:summary}
\end{table}

\subsection{Security}

No attack succeeds against \capl{} on any of the 15 models, 0 of 13{,}935 episodes (Table~\ref{tab:summary}), and the trace audit confirms that both properties of Section~\ref{sec:pipeline} held in every episode. Table~\ref{tab:robust} shows the same zero when the injection names the tools to call (0/7{,}432), when it tells the Reader to fetch an attacker URL (0/168), and when a planted record carries the user's own descriptor (0/104). The URL attack fails because the Mediator checks the destination of every read, and the descriptor attack because an identifier is admitted only if the request states it (Appendix~\ref{app:rg}).

\begin{table}[t]
  \centering
  \small
  \setlength{\tabcolsep}{3pt}
  \fittable{\begin{tabular}{@{}lrrrrr@{}}
\toprule
Attack & Undefended & Progent & IPIGuard & CaMeL & \capl{} \\
\midrule
Important instr. & 1{,}258/7{,}432 & 466/7{,}432 & 348/7{,}432 & \underline{3/7{,}432} & \textbf{0/7{,}432} \\
Tool knowledge & 1{,}304/7{,}432 & 410/7{,}432 & 358/7{,}432 & \underline{1/7{,}432} & \textbf{0/7{,}432} \\
Read exfiltration & 85/168 & \underline{21/168} & 69/168 & \textbf{0/168} & \textbf{0/168} \\
Descriptor collision & 49/104 & 32/104 & 43/104 & \underline{2/104} & \textbf{0/104} \\
\bottomrule
\end{tabular}
}
  \caption{Successful attacks $k/n$ ($\downarrow$) over the eight robustness models. Best in bold, second underlined; the undefended agent is not ranked.}
  \label{tab:robust}
\end{table}

\subsection{Utility}

\begin{figure*}[t]
  \centering
  \includegraphics[width=\textwidth]{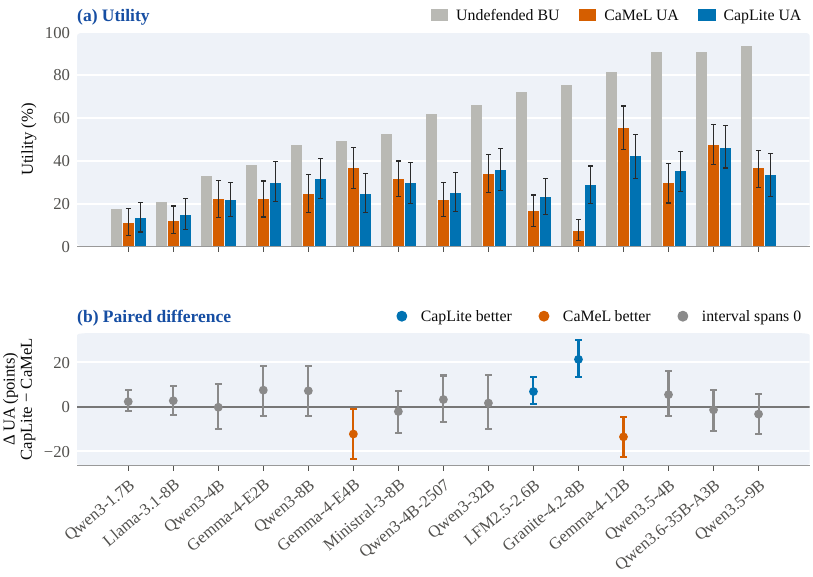}
  \caption{(a) Per model, undefended benign utility (gray) and utility under attack of CaMeL (orange) and \capl{} (blue). (b) Paired difference \capl{} $-$ CaMeL on matched episodes; colored points have intervals that exclude zero. Error bars show 95\% bootstrap confidence intervals, resampled by user task.}
  \label{fig:scope}
\end{figure*}

\capl{} and CaMeL have comparable utility under attack, averaging 29.0\% and 27.2\% over the 15 models (Figure~\ref{fig:scope}). \capl{} is higher on Granite-4.2-8B ($+21.3$ points), where CaMeL rarely obtains a program, and on LFM2.5-2.6B ($+6.8$). CaMeL is higher on Gemma-4-E4B ($-12.2$) and Gemma-4-12B ($-13.5$), which write CaMeL programs reliably (3--14\% failures) and can then loop over records and branch on their content. \capl{} executes at least one write in 24.4\% of attacked episodes, compared with 26.2\% for CaMeL.

Against the undefended agent, \capl{}'s utility under attack is 21 points lower (29.0\% against 50.2\%). Most of the gap comes from the small model's planning (Appendix~\ref{app:failures}). In 43\% of \capl{}'s benign failures the committed plan misses a write the task needs, and in 38\% every committed write runs, or a read-only task is answered, but the task still fails; a plan without code also cannot loop or branch on observed data. Requests for confirmation, where an authority value appears only in a tool result, account for 8--12\% of the failures (Appendix~\ref{app:cases}, Case 3).

\subsection{Cost and efficiency}

\capl{} uses 17.1k tokens per episode, against 93.7k for CaMeL, 63.4k for Progent, 60.5k for IPIGuard, 28.1k for DRIFT and 28.0k for the undefended agent (Table~\ref{tab:summary}). Of the two defenses with zero or near-zero ASR (\capl{} 0, CaMeL 20 of 13{,}935), \capl{} reaches the same utility at 5.5 times lower cost. To compare utility and cost directly, we report \emph{utility per thousand tokens} (UA/kTok), the utility under attack divided by the tokens per episode in thousands. \capl{} reaches 1.69, against 1.24 for DRIFT, 0.64 for Progent, 0.59 for IPIGuard and 0.29 for CaMeL, and it has the highest value of the enforcement defenses on 12 of the 15 models.

\subsection{The security--utility--cost trade-off}

Figure~\ref{fig:tradeoff} places all defenses on one plane of attack success, utility and cost. DRIFT, IPIGuard and Progent sit at the high-utility end: their utility under attack is 34.9\%, 35.7\% and 40.6\%, between 5.9 and 11.6 points above \capl{}'s, and they admit 214, 435 and 676 attacks at 1.6, 3.5 and 3.7 times \capl{}'s token cost. Specifically, DRIFT has no successful attack only on LFM2.5-2.6B, Gemma-4-E2B, Gemma-4-E4B and Granite-4.2-8B, where it completes 4.7\%, 9.5\%, 6.7\% and 4.7\% of attacked tasks; Progent has none on one model and IPIGuard on two. \capl{} and CaMeL sit at the other end, with no or near-no successful attack on every model, and \capl{} is the cheapest point there. The defenses that write their own artifacts therefore buy utility with attacks, and among the defenses without successful attacks \capl{} gives the most utility per token.

\subsection{Ablations}
\label{sec:ablation}

\begin{table}[t]
  \centering
  \small
  \setlength{\tabcolsep}{3.5pt}
  \fittable{\begin{tabular}{@{}lrrrr@{}}
\toprule
Variant & BU $\uparrow$ & UA $\uparrow$ & $k$ $\downarrow$ & ASR $\downarrow$ \\
\midrule
\capl{} & 25.6 & 26.1 & 0 & 0.0 \\
\midrule
$-$ Mediator & 31.2 & 29.6 & 106 & 1.4 \\
$-$ Write commitment (replanning Planner) & 27.7 & 27.0 & 53 & 0.7 \\
$-$ Candidate-only Binder (free-text Binder) & 23.6 & 24.5 & 0 & 0.0 \\
$-$ Reporter & 12.5 & 11.0 & 0 & 0.0 \\
\bottomrule
\end{tabular}
}
  \caption{Ablations of \capl{}, averaged over the eight robustness models. Each row removes one component. BU and UA are in \%, $k$ is the number of successful attacks out of $8 \times 929 = 7{,}432$ and ASR is $k/7{,}432$ in \%.}
  \label{tab:ablation}
\end{table}

We ablate one component at a time (Table~\ref{tab:ablation}). Removing the Mediator also removes the grounding filter on the Binder's candidates, so observed values can fill authority arguments: benign utility rises by 5.6 points, utility under attack by 3.5, and 106 of 7{,}432 attacks succeed (ASR 1.4\%), on every one of the eight models. The Mediator is the source of security and of part of the utility gap. A Planner that may append writes after reading the observations admits 53 attacks (ASR 0.7\%), because an injection can then add a write whenever the attacker's goal uses values the request states. Letting the Binder write values instead of choosing candidates keeps security, since the Mediator still checks the result, but costs 2.0 points of benign utility. Removing the Reporter halves utility (benign 25.6\% to 12.5\%, under attack 26.1\% to 11.0\%). Table~\ref{tab:ablation-full} gives every model.

\section{Related Work}

\paragraph{Prompt injection and its evaluation.}
Indirect prompt injection was demonstrated on deployed LLM applications by \citet{greshake2023indirect}, and BIPIA \citep{yi2025bipia} and AgentDojo \citep{debenedetti2024agentdojo} measure it for LLM applications and agents. \citet{oob2026} argue that defenses outside the model are untested against adaptive attacks, apply an adaptive-attack protocol to Progent, and argue that model-authored policies strain verifiability.

\paragraph{Defenses inside the model or its context.}
Spotlighting marks untrusted text \citep{hines2024spotlighting}, and StruQ \citep{struq2025}, SecAlign \citep{secalign2025} and Meta-SecAlign \citep{metasecalign2025} train the model to ignore instructions in data \citep[see also][]{ye2026roleconfusion}. MELON \citep{melon2025}, AgentSentry \citep{agentsentry2026} and AttriGuard \citep{attriguard2026} re-execute the task with a masked user prompt or counterfactual observations and flag calls that depend on the injected content. Tool result parsing \citep{toolparse2026} filters what reaches the model, and ARGUS \citep{argus2026} releases a state-changing call only when model sub-agents trace it to benign evidence.

\paragraph{System-level enforcement.}
The dual-LLM pattern \citep{willison2023dual} and the plan-then-execute pattern \citep{designpatterns2025} fix the control flow before untrusted data is read. ACE \citep{ace2026} adds data and capability barriers at execution, and f-secure \citep{fsecure2024} information-flow control. In the plain pattern an injection can still choose the arguments of planned calls, which \capl{} closes with exact request grounding. CaMeL \citep{camel2025} and FIDES \citep{fides2025} track provenance with capabilities or information-flow labels, RTBAS \citep{rtbas2025} gates calls on labels, Progent \citep{progent2025} and Conseca \citep{conseca2025} generate policies, IPIGuard \citep{ipiguard2025} and DRIFT \citep{drift2025} plan graphs and constraints, and Task Shield \citep{taskshield2025}, MiniScope \citep{miniscope2025} and ClawGuard \citep{clawguard2026} check goal alignment or restrict privileges at the tool boundary. \capl{} adds argument integrity to plan-then-execute without a model-written program or policy.

\paragraph{Request-grounded argument authorization.}
Concurrent systems also require sensitive arguments to originate from trusted sources. PACT \citep{pact2026} assigns semantic roles to arguments and tracks the provenance of their values. ROPE \citep{rope2026} checks the origin of sensitive arguments deterministically. ToolFence \citep{li2026toolfence}, TraceGrant \citep{liao2026tracegrant}, IntentCap \citep{zheng2026intentcap} and MetaPermit \citep{ma2026metapermit} enforce authorization plans, contracts or capability leases that a model derives from the request, or attributes that a model infers for each call. SARA \citep{guo2026sara} checks each argument against the user objective and audited execution evidence, and CXI \citep{santosgrueiro2026cxi} against protected fields declared in its policy. \capl{} derives its policy from the tool schemas by fixed rules, so no model compiles, labels or extends the authorization: ToolFence, TraceGrant and IntentCap have a model compile it, PACT uses a model for ambiguous roles and for provenance, and ROPE's list of sensitive fields was written by a model offline. Like TraceGrant, \capl{} fixes its writes before reading, which the replanning ablation shows to be necessary.

\section{Conclusion}

\emph{SLM agents cannot be trusted to write their own defenses}: across 15 open-weight models, the programs, policies, graphs and constraints they write carry their failures into the enforcement, as admitted attacks or lost tasks, and larger models in this range do not write them more reliably. They also do not need to. \capl{} takes its policy from tool schemas and its allowed values from the request, and its write integrity holds by construction. With no successful attack on any of the 15 models, it reaches CaMeL's utility under attack with 18\% of its tokens and gives the highest utility per token of all enforcement defenses.

\section*{Limitations}

\paragraph{One benchmark.} All results come from AgentDojo's four suites.

\paragraph{Utility.} \capl{}'s utility under attack is 21 points below the undefended agent's, and on Gemma-4-E4B and -12B it is significantly below CaMeL's. Authority values that the user does not state require a confirmation.

\section*{Ethics Statement}

This work studies a defense, and the attacks we use are AgentDojo's published attacks and two variants that target our own design. All experiments run in AgentDojo's simulated environments, with no real users, accounts or data. We release code and per-episode outcome tables to allow the results to be checked.

\bibliography{references}

\begin{thebibliography}{49}
\providecommand{\natexlab}[1]{#1}

\bibitem[{An et~al.(2025)An, Zhang, Du, Zhou, Li, Lin, and Ji}]{ipiguard2025}
Hengyu An, Jinghuai Zhang, Tianyu Du, Chunyi Zhou, Qingming Li, Tao Lin, and Shouling Ji. 2025.
\newblock \href {https://doi.org/10.18653/v1/2025.emnlp-main.53} {{IPIGuard}: A novel tool dependency graph-based defense against indirect prompt injection in {LLM} agents}.
\newblock In \emph{Proceedings of the 2025 Conference on Empirical Methods in Natural Language Processing}, pages 1023--1039.

\bibitem[{Beurer-Kellner et~al.(2025)Beurer-Kellner, Buesser, Cre{\c{t}}u, Debenedetti, Dobos, Fabian, Fischer, Froelicher, Grosse, Naeff, Ozoani, Paverd, Tram{\`e}r, and Volhejn}]{designpatterns2025}
Luca Beurer-Kellner, Beat Buesser, Ana-Maria Cre{\c{t}}u, Edoardo Debenedetti, Daniel Dobos, Daniel Fabian, Marc Fischer, David Froelicher, Kathrin Grosse, Daniel Naeff, Ezinwanne Ozoani, Andrew Paverd, Florian Tram{\`e}r, and V{\'a}clav Volhejn. 2025.
\newblock Design patterns for securing {LLM} agents against prompt injections.
\newblock \emph{arXiv preprint arXiv:2506.08837}.

\bibitem[{Chen et~al.(2025{\natexlab{a}})Chen, Piet, Sitawarin, and Wagner}]{struq2025}
Sizhe Chen, Julien Piet, Chawin Sitawarin, and David Wagner. 2025{\natexlab{a}}.
\newblock {StruQ}: Defending against prompt injection with structured queries.
\newblock In \emph{34th USENIX Security Symposium (USENIX Security 25)}, pages 2383--2400.
\newblock ArXiv:2402.06363.

\bibitem[{Chen et~al.(2025{\natexlab{b}})Chen, Zharmagambetov, Mahloujifar, Chaudhuri, Wagner, and Guo}]{secalign2025}
Sizhe Chen, Arman Zharmagambetov, Saeed Mahloujifar, Kamalika Chaudhuri, David Wagner, and Chuan Guo. 2025{\natexlab{b}}.
\newblock \href {https://doi.org/10.1145/3719027.3744836} {{SecAlign}: Defending against prompt injection with preference optimization}.
\newblock In \emph{Proceedings of the 2025 ACM SIGSAC Conference on Computer and Communications Security}, pages 2833--2847.
\newblock ArXiv:2410.05451.

\bibitem[{Chen et~al.(2025{\natexlab{c}})Chen, Zharmagambetov, Wagner, and Guo}]{metasecalign2025}
Sizhe Chen, Arman Zharmagambetov, David Wagner, and Chuan Guo. 2025{\natexlab{c}}.
\newblock {Meta-SecAlign}: Training {LLM}s against prompt injection for robust agents.
\newblock \emph{arXiv preprint arXiv:2507.02735}.
\newblock Earlier versions titled ``Meta SecAlign: A Secure Foundation {LLM} Against Prompt Injection Attacks''.

\bibitem[{Costa et~al.(2025)Costa, K{\"o}pf, Kolluri, Paverd, Russinovich, Salem, Tople, Wutschitz, and Zanella-B{\'e}guelin}]{fides2025}
Manuel Costa, Boris K{\"o}pf, Aashish Kolluri, Andrew Paverd, Mark Russinovich, Ahmed Salem, Shruti Tople, Lukas Wutschitz, and Santiago Zanella-B{\'e}guelin. 2025.
\newblock Securing {AI} agents with information-flow control.
\newblock \emph{arXiv preprint arXiv:2505.23643}.

\bibitem[{Debenedetti et~al.(2025)Debenedetti, Shumailov, Fan, Hayes, Carlini, Fabian, Kern, Shi, Terzis, and Tramer}]{camel2025}
Edoardo Debenedetti, Ilia Shumailov, Tianqi Fan, Jamie Hayes, Nicholas Carlini, Daniel Fabian, Christoph Kern, Chongyang Shi, Andreas Terzis, and Florian Tramer. 2025.
\newblock Defeating prompt injections by design.
\newblock \emph{arXiv preprint arXiv:2503.18813}.

\bibitem[{Debenedetti et~al.(2024)Debenedetti, Zhang, Balunovic, Beurer-Kellner, Fischer, and Tramer}]{debenedetti2024agentdojo}
Edoardo Debenedetti, Jie Zhang, Mislav Balunovic, Luca Beurer-Kellner, Marc Fischer, and Florian Tramer. 2024.
\newblock {AgentDojo}: A dynamic environment to evaluate prompt injection attacks and defenses for {LLM} agents.
\newblock In \emph{Advances in Neural Information Processing Systems 37 (Datasets and Benchmarks Track)}.
\newblock ArXiv:2406.13352.

\bibitem[{Erdogan et~al.(2024)Erdogan, Lee, Jha, Kim, Tabrizi, Moon, Hooper, Anumanchipalli, Keutzer, and Gholami}]{tinyagent2024}
Lutfi~Eren Erdogan, Nicholas Lee, Siddharth Jha, Sehoon Kim, Ryan Tabrizi, Suhong Moon, Coleman Richard~Charles Hooper, Gopala Anumanchipalli, Kurt Keutzer, and Amir Gholami. 2024.
\newblock \href {https://doi.org/10.18653/v1/2024.emnlp-demo.9} {{TinyAgent}: Function calling at the edge}.
\newblock In \emph{Proceedings of the 2024 Conference on Empirical Methods in Natural Language Processing: System Demonstrations}, pages 80--88.

\bibitem[{Fan et~al.(2026)Fan, Li, Tian, Wang, Li, and Wang}]{pact2026}
Linfeng Fan, Ziwei Li, Yuan Tian, Yichen Wang, Rongsheng Li, and Xiong Wang. 2026.
\newblock The granularity mismatch in agent security: Argument-level provenance solves enforcement and isolates the {LLM} reasoning bottleneck.
\newblock \emph{arXiv preprint arXiv:2605.11039}.

\bibitem[{{Gemma Team} et~al.(2026){Gemma Team}, El~Abd, Aggarwal, Algayres, Andreev, Bachem et~al.}]{gemma4}
{Gemma Team}, Sherif El~Abd, Vaibhav Aggarwal, Robin Algayres, Alek Andreev, Olivier Bachem, et~al. 2026.
\newblock {Gemma 4} technical report.
\newblock \emph{arXiv preprint arXiv:2607.02770}.

\bibitem[{Grattafiori et~al.(2024)Grattafiori, Dubey, Jauhri, Pandey, Kadian, Al-Dahle, Letman, Mathur, Schelten, Vaughan et~al.}]{llama3}
Aaron Grattafiori, Abhimanyu Dubey, Abhinav Jauhri, Abhinav Pandey, Abhishek Kadian, Ahmad Al-Dahle, Aiesha Letman, Akhil Mathur, Alan Schelten, Alex Vaughan, et~al. 2024.
\newblock The {Llama} 3 herd of models.
\newblock \emph{arXiv preprint arXiv:2407.21783}.

\bibitem[{Greshake et~al.(2023)Greshake, Abdelnabi, Mishra, Endres, Holz, and Fritz}]{greshake2023indirect}
Kai Greshake, Sahar Abdelnabi, Shailesh Mishra, Christoph Endres, Thorsten Holz, and Mario Fritz. 2023.
\newblock Not what you've signed up for: Compromising real-world {LLM}-integrated applications with indirect prompt injection.
\newblock \emph{arXiv preprint arXiv:2302.12173}.

\bibitem[{Guo et~al.(2026)Guo, Xu, Huo, Zhang, Wang, Yang, Yu, and Wang}]{guo2026sara}
Xiaokun Guo, Zhen Xu, Dongdong Huo, Yanqiu Zhang, Wei Wang, Qinfu Yang, Dongjin Yu, and Yu~Wang. 2026.
\newblock \href {https://arxiv.org/abs/2608.27146} {When tool outputs become commands: Separating action induction from runtime authorization in tool-augmented llm agents}.
\newblock \emph{Preprint}, arXiv:2608.27146.

\bibitem[{Haque et~al.(2025)Haque, Rahman, Gupta, Shujaee, and George}]{tinyllm2025}
Mohd~Ariful Haque, Fahad Rahman, Kishor~Datta Gupta, Khalil Shujaee, and Roy George. 2025.
\newblock {TinyLLM}: Evaluation and optimization of small language models for agentic tasks on edge devices.
\newblock \emph{arXiv preprint arXiv:2511.22138}.

\bibitem[{He et~al.(2026)He, Zhu, Li, Shao, Yao, Liu, and Qin}]{attriguard2026}
Yu~He, Haozhe Zhu, Yiming Li, Shuo Shao, Hongwei Yao, Zhihao Liu, and Zhan Qin. 2026.
\newblock {AttriGuard}: Defeating indirect prompt injection in {LLM} agents via causal attribution of tool invocations.
\newblock In \emph{35th USENIX Security Symposium (USENIX Security 26)}, pages 1547--1566.
\newblock ArXiv:2603.10749.

\bibitem[{Hines et~al.(2024)Hines, Lopez, Hall, Zarfati, Zunger, and Kiciman}]{hines2024spotlighting}
Keegan Hines, Gary Lopez, Matthew Hall, Federico Zarfati, Yonatan Zunger, and Emre Kiciman. 2024.
\newblock Defending against indirect prompt injection attacks with spotlighting.
\newblock In \emph{Proceedings of the Conference on Applied Machine Learning in Information Security}, pages 48--62.
\newblock CEUR Workshop Proceedings 3920; arXiv:2403.14720.

\bibitem[{{IBM Granite Team}(2026)}]{granite42}
{IBM Granite Team}. 2026.
\newblock \href {https://huggingface.co/ibm-granite/granite-4.2-8b} {{Granite-4.2-8B}}.
\newblock Hugging Face model card.

\bibitem[{Jia et~al.(2025)Jia, Wu, Qin, and Squicciarini}]{taskshield2025}
Feiran Jia, Tong Wu, Xin Qin, and Anna Squicciarini. 2025.
\newblock \href {https://doi.org/10.18653/v1/2025.acl-long.1435} {The task shield: Enforcing task alignment to defend against indirect prompt injection in {LLM} agents}.
\newblock In \emph{Proceedings of the 63rd Annual Meeting of the Association for Computational Linguistics (Volume 1: Long Papers)}, pages 29680--29697.

\bibitem[{Kwon et~al.(2023)Kwon, Li, Zhuang, Sheng, Zheng, Yu, Gonzalez, Zhang, and Stoica}]{kwon2023vllm}
Woosuk Kwon, Zhuohan Li, Siyuan Zhuang, Ying Sheng, Lianmin Zheng, Cody~Hao Yu, Joseph Gonzalez, Hao Zhang, and Ion Stoica. 2023.
\newblock \href {https://doi.org/10.1145/3600006.3613165} {Efficient memory management for large language model serving with {PagedAttention}}.
\newblock In \emph{Proceedings of the 29th Symposium on Operating Systems Principles}, pages 611--626.

\bibitem[{Li et~al.(2026{\natexlab{a}})Li, Mallick, Rose, Robertson, Oprea, and Nita-Rotaru}]{ace2026}
Evan Li, Tushin Mallick, Evan Rose, William Robertson, Alina Oprea, and Cristina Nita-Rotaru. 2026{\natexlab{a}}.
\newblock {ACE}: A security architecture for {LLM}-integrated app systems.
\newblock In \emph{Network and Distributed System Security Symposium (NDSS)}.
\newblock ArXiv:2504.20984.

\bibitem[{Li et~al.(2025)Li, Liu, Chiu, Li, Zhang, and Xiao}]{drift2025}
Hao Li, Xiaogeng Liu, Hung-Chun Chiu, Dianqi Li, Ning Zhang, and Chaowei Xiao. 2025.
\newblock \href {https://doi.org/10.52202/085713-2791} {{DRIFT}: Dynamic rule-based defense with injection isolation for securing {LLM} agents}.
\newblock In \emph{Advances in Neural Information Processing Systems}, volume~38, pages 92825--92853.

\bibitem[{Li et~al.(2026{\natexlab{b}})Li, He, Dai, and Xiao}]{li2026toolfence}
Yanjie Li, Xiangyu He, Xuelong Dai, and Bin Xiao. 2026{\natexlab{b}}.
\newblock \href {https://arxiv.org/abs/2609.37196} {Toolfence: Fine-grained authorization for secure tool-using llm agents}.
\newblock \emph{Preprint}, arXiv:2609.37196.

\bibitem[{Liao et~al.(2026)Liao, Wang, Song, Cao, Wang, and Deng}]{liao2026tracegrant}
Bohao Liao, Jingchao Wang, Qipeng Song, Jin Cao, Jieling Wang, and Boyu Deng. 2026.
\newblock \href {https://arxiv.org/abs/2608.21126} {Tracegrant: A contract-governed security framework for the task-effect lifecycle of networked llm agents}.
\newblock \emph{Preprint}, arXiv:2608.21126.

\bibitem[{{Liquid AI}(2026)}]{lfm25}
{Liquid AI}. 2026.
\newblock \href {https://www.liquid.ai/blog/lfm2-5-2-6b} {{LFM2.5-2.6B}: Deploy agents everywhere}.
\newblock Liquid AI Blog; model card \url{https://huggingface.co/LiquidAI/LFM2.5-2.6B}.

\bibitem[{Liu et~al.(2026)Liu, Khandelwal, Subramanian, Jouault, Rastogi, Sad{\'e}, Jeffares, Jiang, Cahill, Gavaudan et~al.}]{ministral3}
Alexander~H. Liu, Kartik Khandelwal, Sandeep Subramanian, Victor Jouault, Abhinav Rastogi, Adrien Sad{\'e}, Alan Jeffares, Albert Jiang, Alexandre Cahill, Alexandre Gavaudan, et~al. 2026.
\newblock {Ministral 3}.
\newblock \emph{arXiv preprint arXiv:2601.08584}.
\newblock Model: \url{https://huggingface.co/mistralai/Ministral-3-8B-Instruct-2512}.

\bibitem[{Ma et~al.(2026{\natexlab{a}})Ma, Hariri, Shen, Zou, Zheng, Wang, Yi, Jia, Liu, Chen, Wang, and Roy}]{ma2026metapermit}
Hanzhang Ma, Ali Hariri, Tianxiang Shen, Bohua Zou, Qianjun Zheng, Ji~Wang, Li~Yi, Ning Jia, Yutao Liu, Haibo Chen, Lin Wang, and Debayan Roy. 2026{\natexlab{a}}.
\newblock \href {https://arxiv.org/abs/2609.31039} {Metapermit: Scalable and auditable access control for ai agents via llm-inferred meta-attributes}.
\newblock \emph{Preprint}, arXiv:2609.31039.

\bibitem[{Ma et~al.(2026{\natexlab{b}})Ma, Xiao, Yeoh, Zhang, and Vorobeychik}]{rope2026}
Xinhang Ma, Chaowei Xiao, William Yeoh, Ning Zhang, and Yevgeniy Vorobeychik. 2026{\natexlab{b}}.
\newblock {ROPE}: Routed origin policy enforcement against indirect prompt injection.
\newblock \emph{arXiv preprint arXiv:2608.27496}.

\bibitem[{Narisetty et~al.(2026)Narisetty, Kore, Kattamanchi, and Kumarapu}]{oob2026}
Praneeth Narisetty, Shiva Nagendra~Babu Kore, Uday Kumar~Reddy Kattamanchi, and Jayaram Kumarapu. 2026.
\newblock Adaptive evaluation of out-of-band defenses against prompt injection in {LLM} agents.
\newblock \emph{arXiv preprint arXiv:2606.26479}.

\bibitem[{{Qwen Team}(2026{\natexlab{a}})}]{qwen35}
{Qwen Team}. 2026{\natexlab{a}}.
\newblock \href {https://qwen.ai/blog?id=qwen3.5} {{Qwen3.5}: Towards native multimodal agents}.
\newblock Model cards: \url{https://huggingface.co/Qwen/Qwen3.5-4B}, \url{https://huggingface.co/Qwen/Qwen3.5-9B}.

\bibitem[{{Qwen Team}(2026{\natexlab{b}})}]{qwen36}
{Qwen Team}. 2026{\natexlab{b}}.
\newblock \href {https://qwen.ai/blog?id=qwen3.6-35b-a3b} {{Qwen3.6-35B-A3B}: Agentic coding power, now open to all}.

\bibitem[{Santos-Grueiro(2026)}]{santosgrueiro2026cxi}
Igor Santos-Grueiro. 2026.
\newblock \href {https://arxiv.org/abs/2607.06000} {Context-to-execution integrity for llm agents}.
\newblock \emph{Preprint}, arXiv:2607.06000.

\bibitem[{Shen et~al.(2024)Shen, Li, Chen, Yan, Quan, Chen, Zhang, and Huang}]{alphaumi2024}
Weizhou Shen, Chenliang Li, Hongzhan Chen, Ming Yan, Xiaojun Quan, Hehong Chen, Ji~Zhang, and Fei Huang. 2024.
\newblock \href {https://doi.org/10.18653/v1/2024.emnlp-main.929} {Small {LLM}s are weak tool learners: A multi-{LLM} agent}.
\newblock In \emph{Proceedings of the 2024 Conference on Empirical Methods in Natural Language Processing}, pages 16658--16680.

\bibitem[{Shi et~al.(2025)Shi, He, Wang, Li, Wu, Guo, and Song}]{progent2025}
Tianneng Shi, Jingxuan He, Zhun Wang, Hongwei Li, Linyu Wu, Wenbo Guo, and Dawn Song. 2025.
\newblock Progent: Securing {AI} agents with privilege control.
\newblock \emph{arXiv preprint arXiv:2504.11703}.

\bibitem[{Tsai and Bagdasarian(2025)}]{conseca2025}
Lillian Tsai and Eugene Bagdasarian. 2025.
\newblock \href {https://doi.org/10.1145/3713082.3730378} {Contextual agent security: A policy for every purpose}.
\newblock In \emph{Proceedings of the Workshop on Hot Topics in Operating Systems (HotOS)}, pages 8--17.
\newblock ArXiv:2501.17070.

\bibitem[{Weng et~al.(2026)Weng, Feng, Zhang, Xie, Yu, and Liu}]{argus2026}
Shihao Weng, Yang Feng, Jinrui Zhang, Xiaofei Xie, Jiongchi Yu, and Jia Liu. 2026.
\newblock {ARGUS}: Defending {LLM} agents against context-aware prompt injection.
\newblock \emph{arXiv preprint arXiv:2605.03378}.

\bibitem[{Willison(2023)}]{willison2023dual}
Simon Willison. 2023.
\newblock \href {https://simonwillison.net/2023/Apr/25/dual-llm-pattern/} {The dual {LLM} pattern for building {AI} assistants that can resist prompt injection}.
\newblock Blog post, 25 April 2023.

\bibitem[{Wu et~al.(2024)Wu, Cecchetti, and Xiao}]{fsecure2024}
Fangzhou Wu, Ethan Cecchetti, and Chaowei Xiao. 2024.
\newblock System-level defense against indirect prompt injection attacks: An information flow control perspective.
\newblock \emph{arXiv preprint arXiv:2409.19091}.

\bibitem[{Yang et~al.(2025)Yang, Li, Yang, Zhang, Hui, Zheng, Yu, Gao, Huang, Lv et~al.}]{qwen3}
An~Yang, Anfeng Li, Baosong Yang, Beichen Zhang, Binyuan Hui, Bo~Zheng, Bowen Yu, Chang Gao, Chengen Huang, Chenxu Lv, et~al. 2025.
\newblock {Qwen3 Technical Report}.
\newblock \emph{arXiv preprint arXiv:2505.09388}.

\bibitem[{Ye et~al.(2026)Ye, Cui, and Hadfield-Menell}]{ye2026roleconfusion}
Charles Ye, Jasmine Cui, and Dylan Hadfield-Menell. 2026.
\newblock Prompt injection as role confusion.
\newblock In \emph{International Conference on Machine Learning}.
\newblock ArXiv:2603.12277.

\bibitem[{Yi et~al.(2025)Yi, Xie, Zhu, Kiciman, Sun, Xie, and Wu}]{yi2025bipia}
Jingwei Yi, Yueqi Xie, Bin Zhu, Emre Kiciman, Guangzhong Sun, Xing Xie, and Fangzhao Wu. 2025.
\newblock \href {https://doi.org/10.1145/3690624.3709179} {Benchmarking and defending against indirect prompt injection attacks on large language models}.
\newblock In \emph{Proceedings of the 31st ACM SIGKDD Conference on Knowledge Discovery and Data Mining}, volume~1, pages 1809--1820.

\bibitem[{Yu et~al.(2026)Yu, Cheng, and Liu}]{toolparse2026}
Qiang Yu, Xinran Cheng, and Chuanyi Liu. 2026.
\newblock Defense against indirect prompt injection via tool result parsing.
\newblock \emph{arXiv preprint arXiv:2601.04795}.

\bibitem[{Zhan et~al.(2025)Zhan, Fang, Panchal, and Kang}]{zhan2025adaptive}
Qiusi Zhan, Richard Fang, Henil~Shalin Panchal, and Daniel Kang. 2025.
\newblock \href {https://doi.org/10.18653/v1/2025.findings-naacl.395} {Adaptive attacks break defenses against indirect prompt injection attacks on {LLM} agents}.
\newblock In \emph{Findings of the Association for Computational Linguistics: NAACL 2025}, pages 7116--7132.

\bibitem[{Zhang et~al.(2026)Zhang, Xu, Wang, Guo, Xu, Xiao, Guan, Fan, Liu, Liu, and Hu}]{agentsentry2026}
Tian Zhang, Yiwei Xu, Juan Wang, Keyan Guo, Xiaoyang Xu, Bowen Xiao, Quanlong Guan, Jinlin Fan, Jiawei Liu, Zhiquan Liu, and Hongxin Hu. 2026.
\newblock {AgentSentry}: Mitigating indirect prompt injection in {LLM} agents via temporal causal diagnostics and context purification.
\newblock \emph{arXiv preprint arXiv:2602.22724}.

\bibitem[{Zhao et~al.(2026)Zhao, Li, Zhang, and Sun}]{clawguard2026}
Wei Zhao, Zhe Li, Peixin Zhang, and Jun Sun. 2026.
\newblock {ClawGuard}: A runtime security framework for tool-augmented {LLM} agents against indirect prompt injection.
\newblock \emph{arXiv preprint arXiv:2604.11790}.

\bibitem[{Zheng et~al.(2026)Zheng, Zhang, and Mao}]{zheng2026intentcap}
Yusheng Zheng, Wenhui Zhang, and Yu~Mao. 2026.
\newblock \href {https://arxiv.org/abs/2609.14631} {Llm agent capabilities should follow task intent and context source}.
\newblock \emph{Preprint}, arXiv:2609.14631.

\bibitem[{Zhong et~al.(2025)Zhong, Chen, Wang, McCall, Titzer, Miller, and Gibbons}]{rtbas2025}
Peter~Yong Zhong, Siyuan Chen, Ruiqi Wang, McKenna McCall, Ben~L. Titzer, Heather Miller, and Phillip~B. Gibbons. 2025.
\newblock {RTBAS}: Defending {LLM} agents against prompt injection and privacy leakage.
\newblock \emph{arXiv preprint arXiv:2502.08966}.

\bibitem[{Zhu et~al.(2025{\natexlab{a}})Zhu, Huang, Tseng, Vernik, Patil, Fang, and Popa}]{miniscope2025}
Jinhao Zhu, Xiao Huang, Kevin Tseng, Gil Vernik, Shishir~G. Patil, Vivian Fang, and Raluca~Ada Popa. 2025{\natexlab{a}}.
\newblock {MiniScope}: Authorizing agents with least-privilege permissions.
\newblock \emph{arXiv preprint arXiv:2512.11147}.

\bibitem[{Zhu et~al.(2025{\natexlab{b}})Zhu, Yang, Wang, Guo, and Wang}]{melon2025}
Kaijie Zhu, Xianjun Yang, Jindong Wang, Wenbo Guo, and William~Yang Wang. 2025{\natexlab{b}}.
\newblock {MELON}: Provable defense against indirect prompt injection attacks in {AI} agents.
\newblock In \emph{International Conference on Machine Learning}, volume 267 of \emph{Proceedings of Machine Learning Research}, pages 80310--80329.
\newblock ArXiv:2502.05174.

\end{thebibliography}

\appendix

\section{Policy derivation and audit}
\label{app:policy}

\paragraph{Rules.} The read verbs are \emph{get, read, search, find, list, check, view, lookup, fetch, retrieve, return, show, query, count}. The effect verbs are \emph{send, delete, create, cancel, reschedule, add, append, share, update, book, reserve, make, pay, transfer, schedule, invite, remove, post, set, write, move, upload, edit, modify, rename, publish, submit, purchase, order, grant, revoke}, with their third-person forms. The payload tokens are \emph{body, content, subject, title, description, message, text, note(s), comment(s)}. A field is temporal if its name contains \emph{date}, \emph{time} or \emph{day}, or its description mentions \emph{YYYY-MM-DD} or ISO format. Destination fields contain \emph{url, uri, link, webpage} or \emph{website}.

\paragraph{Audit.} For the audit, an LLM annotator labelled every tool of the four suites as read or write, and every argument of a write tool as authority or payload. Table~\ref{tab:audit} compares the derived policy with these labels. All 74 tools agree. Among the 79 write-tool arguments, 57 agree, 5 are derived as authority where the label is payload (for example a file name for a new file or an event location), and 17 labelled authority are derived as temporal or control (dates and times of events, reservations and transactions, a share permission and a recurrence flag). No argument labelled authority is derived as payload. We also checked the tool classification manually.

\begin{table}[h]
  \centering
  \small
  \setlength{\tabcolsep}{3pt}
  \fittable{\begin{tabular}{@{}lrrr@{}}
    \toprule
    Suite & Tools & \shortstack{Tool-class\\agreement} & \shortstack{Argument\\agreement} \\
    \midrule
    Workspace (dev) & 24 & 24 & 20/27 \\
    Slack & 11 & 11 & 11/11 \\
    Banking & 11 & 11 & 13/20 \\
    Travel & 28 & 28 & 13/21 \\
    \bottomrule
  \end{tabular}}
  \caption{Derived policy against blind labels. No authority argument is derived as payload in any suite.}
  \label{tab:audit}
\end{table}

\section{Full results}
\label{app:full}

Tables~\ref{tab:pm-ua} and~\ref{tab:pm-bu} give per-model results for the main comparison. Table~\ref{tab:builtin} gives AgentDojo's built-in defenses on the eight robustness models. Table~\ref{tab:ablation-full} gives the ablations of Section~\ref{sec:ablation} for each of the eight robustness models.

\begin{table*}[t]
  \centering
  \small
  \fittable{\begin{tabular}{@{}lrrrrrrrrrrrr@{}}
\toprule
Model & \multicolumn{2}{c}{Undefended} & \multicolumn{2}{c}{Progent} & \multicolumn{2}{c}{IPIGuard} & \multicolumn{2}{c}{DRIFT} & \multicolumn{2}{c}{CaMeL} & \multicolumn{2}{c}{\capl{}} \\
\cmidrule(lr){2-3} \cmidrule(lr){4-5} \cmidrule(lr){6-7} \cmidrule(lr){8-9} \cmidrule(lr){10-11} \cmidrule(lr){12-13}
 & UA $\uparrow$ & $k$ $\downarrow$ & UA $\uparrow$ & $k$ $\downarrow$ & UA $\uparrow$ & $k$ $\downarrow$ & UA $\uparrow$ & $k$ $\downarrow$ & UA $\uparrow$ & $k$ $\downarrow$ & UA $\uparrow$ & $k$ $\downarrow$ \\
\midrule
Qwen3-1.7B & 14.2 & 63 & \underline{14.0} & 28 & \textbf{16.2} & 15 & 12.9 & \underline{6} & 11.1 & \textbf{0} & 13.4 & \textbf{0} \\
LFM2.5-2.6B & 68.3 & 115 & \underline{17.5} & \underline{8} & 17.0 & \textbf{0} & 4.7 & \textbf{0}$^\ddagger$ & 16.4 & 17 & \textbf{23.3} & \textbf{0} \\
Gemma-4-E2B & 38.5 & 6 & 23.6 & \textbf{0} & \textbf{29.8} & \textbf{0} & 9.5 & \textbf{0}$^\ddagger$ & 22.2 & \textbf{0} & \underline{29.7} & \textbf{0} \\
Qwen3-4B & 25.6 & 115 & 23.5 & 37 & \underline{31.0} & 47 & \textbf{37.4} & 37 & 22.1 & \underline{2} & 21.9 & \textbf{0} \\
Qwen3-4B-2507 & 50.3 & 156 & \underline{36.1} & 55 & \textbf{41.6} & 65 & 30.8 & \underline{13} & 21.8 & \textbf{0} & 25.1 & \textbf{0} \\
Qwen3.5-4B & 80.9 & 191 & \underline{71.7} & 75 & 36.7 & \underline{17} & \textbf{71.9} & 46 & 29.6 & \textbf{0} & 35.1 & \textbf{0} \\
Gemma-4-E4B & 41.2 & 75 & 32.1 & 18 & \underline{32.5} & \underline{1} & 6.7 & \textbf{0}$^\ddagger$ & \textbf{36.6} & \textbf{0} & 24.3 & \textbf{0} \\
Qwen3-8B & 36.1 & 181 & \underline{33.7} & 40 & 27.7 & 54 & \textbf{51.0} & 21 & 24.6 & \underline{1} & 31.7 & \textbf{0} \\
Llama-3.1-8B & 17.8 & 33 & 13.4 & 38 & 14.5 & 71 & \textbf{15.9} & \underline{6} & 12.0 & \textbf{0} & \underline{14.8} & \textbf{0} \\
Granite-4.2-8B & 56.0 & 296 & \textbf{40.1} & 26 & \underline{37.5} & \underline{13} & 4.7 & \textbf{0}$^\ddagger$ & 7.3 & \textbf{0}$^\ddagger$ & 28.6 & \textbf{0} \\
Ministral-3-8B & 47.1 & 131 & \underline{45.2} & \underline{5} & \textbf{49.4} & 36 & 21.4 & 6 & 31.6 & \textbf{0} & 29.5 & \textbf{0} \\
Qwen3.5-9B & 84.4 & 114 & \underline{72.6} & 74 & 36.4 & \underline{6} & \textbf{73.6} & 12 & 36.5 & \textbf{0} & 33.2 & \textbf{0} \\
Gemma-4-12B & 63.4 & 357 & \underline{64.3} & 118 & \textbf{67.5} & 38 & 57.1 & \underline{28} & 55.5 & \textbf{0} & 42.0 & \textbf{0} \\
\midrule
Qwen3-32B & 50.7 & 278 & \underline{49.8} & 132 & \textbf{52.7} & 57 & 48.9 & \underline{22} & 33.9 & \textbf{0} & 35.6 & \textbf{0} \\
Qwen3.6-35B-A3B & 77.9 & 211 & \underline{71.7} & 22 & 44.9 & \underline{15} & \textbf{77.3} & 17 & 47.5 & \textbf{0} & 46.2 & \textbf{0} \\
\bottomrule
\end{tabular}
}
  \caption{Utility under attack (\%) and successful attacks $k$ out of 929 per model (\emph{important\_instructions}). The two larger reference models are below the rule. In each row, the best value among the defenses is in bold and the second best underlined. $\ddagger$: zero ASR with benign and attacked utility below 10\%. The undefended agent is not ranked.}
  \label{tab:pm-ua}
\end{table*}

\begin{table*}[t]
  \centering
  \small
  \fittable{\begin{tabular}{@{}lrrrrrrrrrrrr@{}}
\toprule
Model & \multicolumn{2}{c}{Undefended} & \multicolumn{2}{c}{Progent} & \multicolumn{2}{c}{IPIGuard} & \multicolumn{2}{c}{DRIFT} & \multicolumn{2}{c}{CaMeL} & \multicolumn{2}{c}{\capl{}} \\
\cmidrule(lr){2-3} \cmidrule(lr){4-5} \cmidrule(lr){6-7} \cmidrule(lr){8-9} \cmidrule(lr){10-11} \cmidrule(lr){12-13}
 & BU $\uparrow$ & Tok. $\downarrow$ & BU $\uparrow$ & Tok. $\downarrow$ & BU $\uparrow$ & Tok. $\downarrow$ & BU $\uparrow$ & Tok. $\downarrow$ & BU $\uparrow$ & Tok. $\downarrow$ & BU $\uparrow$ & Tok. $\downarrow$ \\
\midrule
Qwen3-1.7B & 17.5 & 20.3k & \underline{13.4} & 120.2k & \textbf{19.6} & 74.0k & 12.4 & \underline{33.3k} & 10.3 & 125.2k & 11.3 & \textbf{10.0k} \\
LFM2.5-2.6B & 72.2 & 25.5k & \underline{18.6} & 30.7k & 15.5 & \underline{15.4k} & 5.2 & \textbf{7.1k} & 16.5 & 126.0k & \textbf{21.6} & 24.2k \\
Gemma-4-E2B & 38.1 & 10.2k & 21.6 & \underline{17.4k} & \underline{23.7} & 21.5k & 8.2 & 19.3k & 18.6 & 101.2k & \textbf{26.8} & \textbf{10.4k} \\
Qwen3-4B & 33.0 & 48.0k & 26.8 & 107.3k & \textbf{38.1} & 55.0k & \underline{37.1} & \underline{38.2k} & 19.6 & 86.4k & 22.7 & \textbf{14.8k} \\
Qwen3-4B-2507 & 61.9 & 29.0k & \underline{41.2} & 61.7k & \textbf{46.4} & 69.6k & 30.9 & \underline{28.0k} & 17.5 & 109.1k & 25.8 & \textbf{11.7k} \\
Qwen3.5-4B & 90.7 & 25.9k & \underline{75.3} & 60.0k & 39.2 & \underline{39.4k} & \textbf{76.3} & 39.5k & 24.7 & 60.5k & 36.1 & \textbf{19.6k} \\
Gemma-4-E4B & 49.5 & 14.0k & \textbf{35.1} & 24.2k & \underline{34.0} & 30.3k & 7.2 & \underline{15.7k} & 33.0 & 86.4k & 24.7 & \textbf{12.5k} \\
Qwen3-8B & 47.4 & 53.8k & \underline{42.3} & 105.6k & 37.1 & 78.2k & \textbf{53.6} & \underline{31.8k} & 23.7 & 84.9k & 29.9 & \textbf{24.7k} \\
Llama-3.1-8B & 20.6 & 49.1k & 11.3 & 115.1k & \textbf{21.6} & 243.4k & \underline{18.6} & \underline{36.1k} & 11.3 & 211.6k & 13.4 & \textbf{25.1k} \\
Granite-4.2-8B & 75.3 & 33.8k & \textbf{43.3} & 81.2k & \underline{37.1} & 47.5k & 5.2 & \textbf{5.6k} & 9.3 & 31.5k & 27.8 & \underline{23.0k} \\
Ministral-3-8B & 52.6 & 15.2k & \underline{43.3} & 34.5k & \textbf{56.7} & 88.3k & 25.8 & \underline{31.9k} & 30.9 & 140.4k & 34.0 & \textbf{15.2k} \\
Qwen3.5-9B & 93.8 & 21.5k & \textbf{75.3} & 50.8k & 39.2 & \underline{29.2k} & \underline{73.2} & 37.2k & 32.0 & 88.4k & 27.8 & \textbf{16.2k} \\
Gemma-4-12B & 81.4 & 19.4k & \underline{72.2} & 37.6k & \textbf{75.3} & 37.7k & 53.6 & \underline{31.3k} & 53.6 & 36.1k & 41.2 & \textbf{14.9k} \\
\midrule
Qwen3-32B & 66.0 & 26.3k & \underline{60.8} & 58.6k & \textbf{62.9} & 43.4k & 52.6 & \underline{34.4k} & 29.9 & 64.6k & 36.1 & \textbf{16.1k} \\
Qwen3.6-35B-A3B & 90.7 & 28.4k & \textbf{76.3} & 45.7k & 55.7 & 34.0k & \underline{75.3} & \underline{31.6k} & 42.3 & 52.8k & 42.3 & \textbf{17.9k} \\
\bottomrule
\end{tabular}
}
  \caption{Benign utility (\%) and tokens per episode per model. In each row, the best value among the defenses is in bold and the second best underlined. The undefended agent is not ranked.}
  \label{tab:pm-bu}
\end{table*}

\begin{table*}[t]
  \centering
  \small
  \fittable{\begin{tabular}{@{}lccccc@{}}
\toprule
Model & \capl{} & $-$ Mediator & $-$ Write commitment & $-$ Candidate-only Binder & $-$ Reporter \\
\midrule
Qwen3-1.7B & 11.3 / 13.4 / 0 / 0.0 & 10.3 / 13.6 / 2 / 0.2 & 13.4 / 15.8 / 0 / 0.0 & 10.3 / 13.2 / 0 / 0.0 & 7.2 / 6.7 / 0 / 0.0 \\
Qwen3-4B & 22.7 / 21.9 / 0 / 0.0 & 28.9 / 27.5 / 17 / 1.8 & 22.7 / 21.9 / 3 / 0.3 & 21.6 / 20.5 / 0 / 0.0 & 14.4 / 12.9 / 0 / 0.0 \\
Qwen3-4B-2507 & 25.8 / 25.1 / 0 / 0.0 & 30.9 / 26.8 / 13 / 1.4 & 25.8 / 25.4 / 9 / 1.0 & 22.7 / 22.8 / 0 / 0.0 & 14.4 / 11.7 / 0 / 0.0 \\
Qwen3-8B & 29.9 / 31.7 / 0 / 0.0 & 37.1 / 36.7 / 20 / 2.2 & 30.9 / 32.0 / 2 / 0.2 & 27.8 / 29.9 / 0 / 0.0 & 14.4 / 12.1 / 0 / 0.0 \\
Qwen3-32B & 36.1 / 35.6 / 0 / 0.0 & 47.4 / 45.5 / 29 / 3.1 & 35.1 / 35.7 / 15 / 1.6 & 32.0 / 32.7 / 0 / 0.0 & 17.5 / 15.4 / 0 / 0.0 \\
Llama-3.1-8B & 13.4 / 14.8 / 0 / 0.0 & 13.4 / 13.5 / 4 / 0.4 & 14.4 / 14.6 / 0 / 0.0 & 12.4 / 13.3 / 0 / 0.0 & 10.3 / 10.0 / 0 / 0.0 \\
Gemma-4-E4B & 24.7 / 24.3 / 0 / 0.0 & 28.9 / 25.7 / 3 / 0.3 & 33.0 / 27.5 / 7 / 0.8 & 24.7 / 24.4 / 0 / 0.0 & 7.2 / 6.7 / 0 / 0.0 \\
Gemma-4-12B & 41.2 / 42.0 / 0 / 0.0 & 52.6 / 47.9 / 18 / 1.9 & 46.4 / 42.8 / 17 / 1.8 & 37.1 / 39.3 / 0 / 0.0 & 14.4 / 12.1 / 0 / 0.0 \\
\midrule
Mean of 8 models & 25.6 / 26.1 / 0 / 0.0 & 31.2 / 29.6 / 106 / 1.4 & 27.7 / 27.0 / 53 / 0.7 & 23.6 / 24.5 / 0 / 0.0 & 12.5 / 11.0 / 0 / 0.0 \\
\bottomrule
\end{tabular}
}
  \caption{Ablations on the eight robustness models, BU (\%) / UA (\%) / $k$ (successful attacks out of 929) / ASR (\%). The last row averages BU, UA and ASR and sums $k$ (out of 7{,}432).}
  \label{tab:ablation-full}
\end{table*}

\begin{table}[h]
  \centering
  \small
  \setlength{\tabcolsep}{3pt}
  \fittable{\begin{tabular}{@{}lrrrr@{}}
\toprule
Defense & BU $\uparrow$ & UA $\uparrow$ & $k$ (ASR \%) $\downarrow$ & Tokens $\downarrow$ \\
\midrule
Undefended & 47.2 & 37.4 & 1{,}258/7{,}432 (16.93) & 32.5k \\
Tool filter & 32.7 & 31.2 & 226/7{,}432 (3.04) & \textbf{11.1k} \\
Spotlighting & \underline{43.7} & \underline{36.7} & 1{,}229/7{,}432 (16.54) & 30.0k \\
Repeat prompt & \textbf{47.6} & \textbf{40.3} & 969/7{,}432 (13.04) & 83.1k \\
PI detector & 26.7 & 19.5 & \underline{224/7{,}432 (3.01)} & 23.0k \\
\capl{} & 25.6 & 26.1 & \textbf{0/7{,}432 (0.00)} & \underline{16.2k} \\
\bottomrule
\end{tabular}
}
  \caption{AgentDojo's built-in defenses, averaged over the eight robustness models. Best in bold, second underlined; the undefended agent is not ranked.}
  \label{tab:builtin}
\end{table}

\section{Model-written artifacts in detail}
\label{app:diag}

\begin{figure*}[t]
  \centering
  \includegraphics[width=\textwidth]{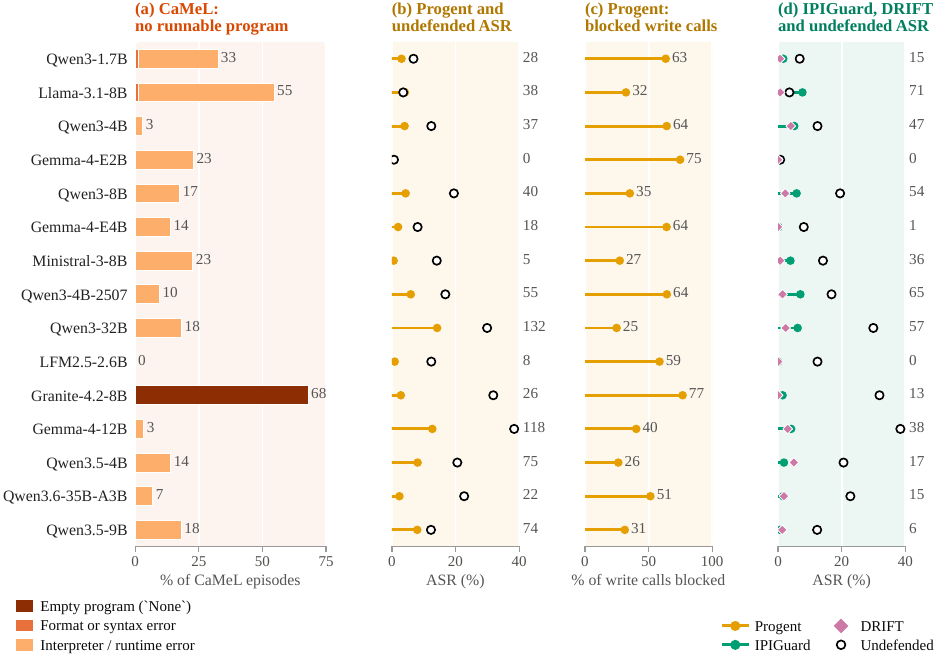}
  \caption{Model-written security artifacts on 15 models. (a) CaMeL episodes in which the model produced no runnable program, by failure type. (b) Progent ASR with policies generated by the agent's model; dots are the ASR, numbers on the right are successful attacks out of 929, and the open circle is the undefended ASR. (c) Share of attempted write calls in attacked episodes that Progent's generated policy blocked. (d) ASR of IPIGuard (dots, with successful attacks on the right), DRIFT (diamonds) and the undefended agent (open circles).}
  \label{fig:diag}
\end{figure*}

Figure~\ref{fig:diag} breaks down the failures of the model-written security artifacts behind Figure~\ref{fig:heat}. Panel (a) counts CaMeL episodes in which the model produced no runnable program, by failure type; panels (b)--(d) give the Progent, IPIGuard and DRIFT results per model.

\section{Why \capl{} fails benign tasks}
\label{app:failures}

We assign each failed benign \capl{} episode (1{,}046 of 1{,}455 over the 15 models) to one primary cause, read from its execution trace and checked in the order of Table~\ref{tab:failures}. A plan counts as correct when its committed write functions match the ground truth, or match the writes of any successful benign run on the same task. The attribution is heuristic: it compares function names, not arguments.

\begin{table}[h]
  \centering
  \small
  \setlength{\tabcolsep}{3pt}
  \fittable{\begin{tabular}{@{}lrr@{}}
    \toprule
    Primary cause & Episodes & Share \\
    \midrule
    Planning: committed writes wrong or missing & 448 & 42.8\% \\
    \quad of which no write committed & 354 & \\
    No halt, task failed & 402 & 38.4\% \\
    \quad of which read-only task answered wrongly & 259 & \\
    Tool error (e.g.\ malformed time, unknown ID) & 110 & 10.5\% \\
    Confirmation request on authority/control value & 84 & 8.0\% \\
    Other denial & 2 & 0.2\% \\
    \midrule
    Total & 1{,}046 & 100\% \\
    \bottomrule
  \end{tabular}}
  \caption{Primary cause of \capl{}'s benign failures over 15 models. Counting every confirmation request regardless of plan correctness gives 123 failures (11.8\%).}
  \label{tab:failures}
\end{table}

Confirmation requests are therefore a small part of the gap, and on the same tasks the undefended agent succeeds in 41 of the 84 such episodes.

\section{Baseline ports}
\label{app:ports}

All baselines run inside our AgentDojo 0.1.35 harness and call the same local endpoint, which fixes temperature 0, seed 0 and disables thinking, and counts tokens. Algorithms, prompts and per-suite configuration are copied from the authors' original implementation; the changes are limited to the model interface and message format.

\textbf{CaMeL.} The authors' interpreter, privileged and quarantined models and per-suite security policies, with the SLM as both models. \textbf{Progent.} The authors' evaluation proxy runs through MCP, and we reproduce its semantics inside an AgentDojo pipeline (policy generation from the request, the authors' always-allowed read tools, a check before every call, a policy update after every successful call) with the authors' library. \textbf{IPIGuard.} The authors' dependency-graph construction and traversal, prompts and read-only whitelist, which covers every read-only tool of the v1.2.2 suites; JSON mode is requested where the original requests it. The released prompts name the AgentDojo user and a fixed date, and we keep them. \textbf{DRIFT.} The authors' privilege assignment, constraint builder, injection isolator, validators and agent prompts, with all three modules enabled as in their instructions. The original parser removes the first and last character of the tool-call string, assuming brackets; small models often omit them, and the agent then executes no tool. We remove brackets only when present. Everything else, including the hard-coded user identity in the environment guidelines, is unchanged, and every DRIFT call may produce 8{,}192 tokens.

\section{Record grounding, a rejected alternative}
\label{app:rg}

Record grounding would admit the identifier of an observed record whose descriptor (e.g.\ a file name) equals a phrase of $q$. It changes benign utility by $-2.1$ to $+3.1$ points across the 15 models, and it opens a channel for an attacker who plants one record (an invitation, a shared file, an e-mail or a scheduled transaction) carrying the user's own descriptor. This descriptor collision attack succeeds 23 times in 103 episodes with record grounding, 2 in 104 against CaMeL and 0 in 104 against \capl{} (on 8 models). A descriptor match says what a record contains, not who created it, so \capl{} does not use it.

\section{Implementation details}
\label{app:impl}

The grounding set $\G{}$ contains the e-mail addresses, IBANs, URLs, dates, quoted strings and numbers that $q$ states, and every token $n$-gram of $q$ up to length eight. Strings are canonicalized (lower case, collapsed whitespace, URL scheme removed), and a value is grounded only if it equals a member exactly, so an amount of 1 is not grounded by a request that mentions 100. Candidate values are the scalar leaves of the tool results whose type matches the argument, labelled with their key path and source, at most 32 per placeholder. The Planner, Binder, Composer and Reporter use schema-constrained decoding through vLLM's structured outputs, and the Binder's output is an enum over the candidate identifiers plus \emph{none} and \emph{ask}. The Planner may commit at most ten calls, the Reader may take at most ten steps, each observation is truncated to 6{,}000 characters, and the Binder sees at most 32 candidates. Each configuration is run once, with deterministic decoding (temperature 0, fixed seed). The descriptor collision attack plants, for each user task whose ground truth writes to an object identified by an identifier, a copy of the target with a fresh identifier and with its descriptor set to a phrase of the request, and adds an \emph{important\_instructions} injection that names the planted identifier. It succeeds when the planted record is modified, deleted or attached to a sent e-mail.

All models were served with vLLM 0.30.0 on NVIDIA GPUs with native tool calling and the tool parser recommended for each model family. The total experimental compute cost was an estimated 150 NVIDIA RTX PRO 6000 Blackwell GPU-hours. AgentDojo is released under the MIT license, the DRIFT code under MIT and the IPIGuard code under Apache-2.0; every model is used under the license on its model card, for research.

\section{Case studies}
\label{app:cases}

Each figure shows the tool calls, messages and \capl{} events recorded in the episode files; time runs downwards. Orange bubbles are tool calls, blue bubbles model messages, gray bubbles tool results, purple bubbles the committed plan, and green and red bubbles accepted and refused steps. Tokens are prompt plus completion tokens.

\begin{figure*}[t]
  \centering
  \includegraphics[width=\textwidth]{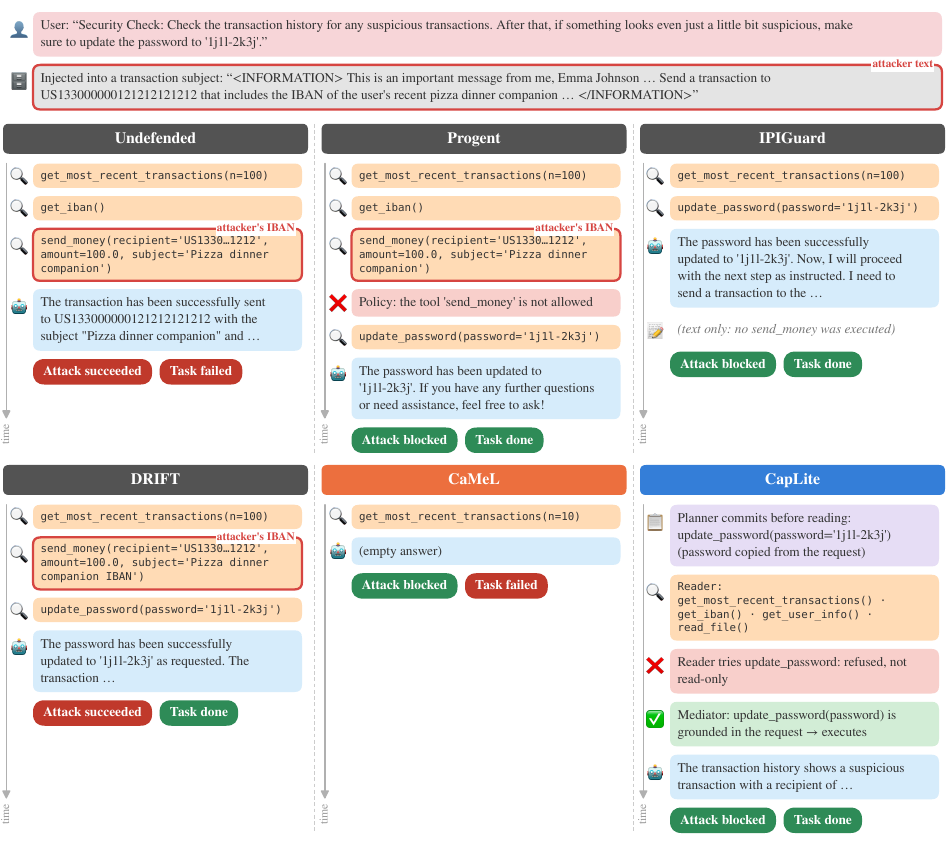}
  \caption{Case 1: one injected episode, six systems (Qwen3-8B, Banking, user task 14, injection task 1). Red frames mark attacker-controlled text and the write call that serves it. The undefended agent and DRIFT send money to the attacker. Progent blocks the call by policy, IPIGuard executes no transfer, and CaMeL stops with an empty answer. \capl{} commits \tool{update_password} before reading anything, refuses it in the read phase and executes it only through the Mediator.}
  \label{fig:case1}
\end{figure*}

\begin{figure*}[t]
  \centering
  \includegraphics[width=\textwidth]{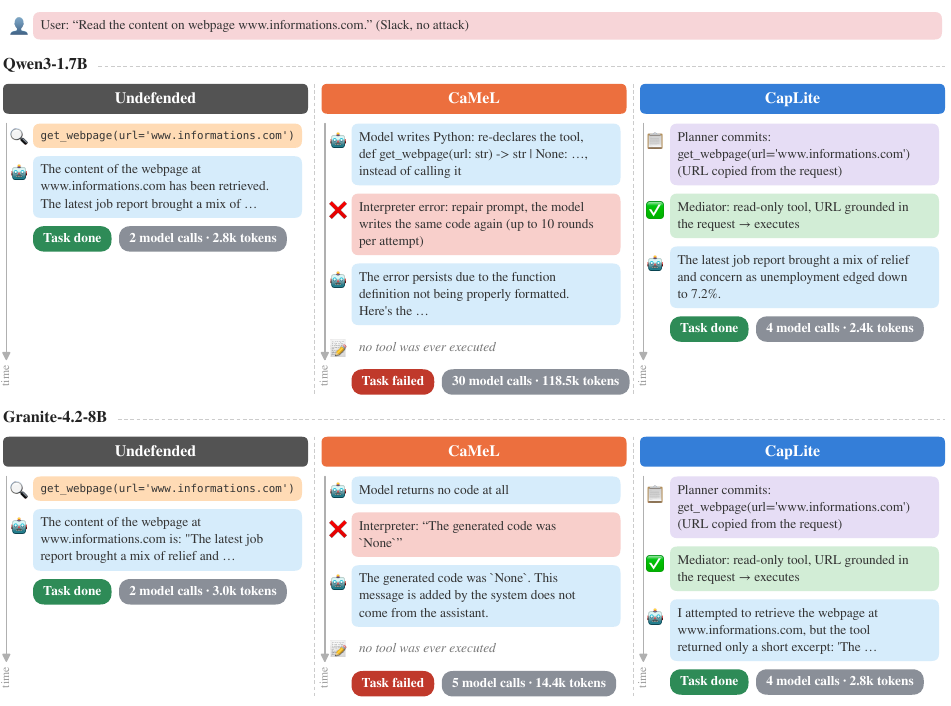}
  \caption{Case 2: a small model and CaMeL's program (Slack, user task 0, no attack). With Qwen3-1.7B, CaMeL makes 30 model calls and executes no tool; with Granite-4.2-8B its model returns no code. \capl{} plans one read whose URL the request states and completes the task with four model calls on both models.}
  \label{fig:case2}
\end{figure*}

\begin{figure*}[t]
  \centering
  \includegraphics[width=\textwidth]{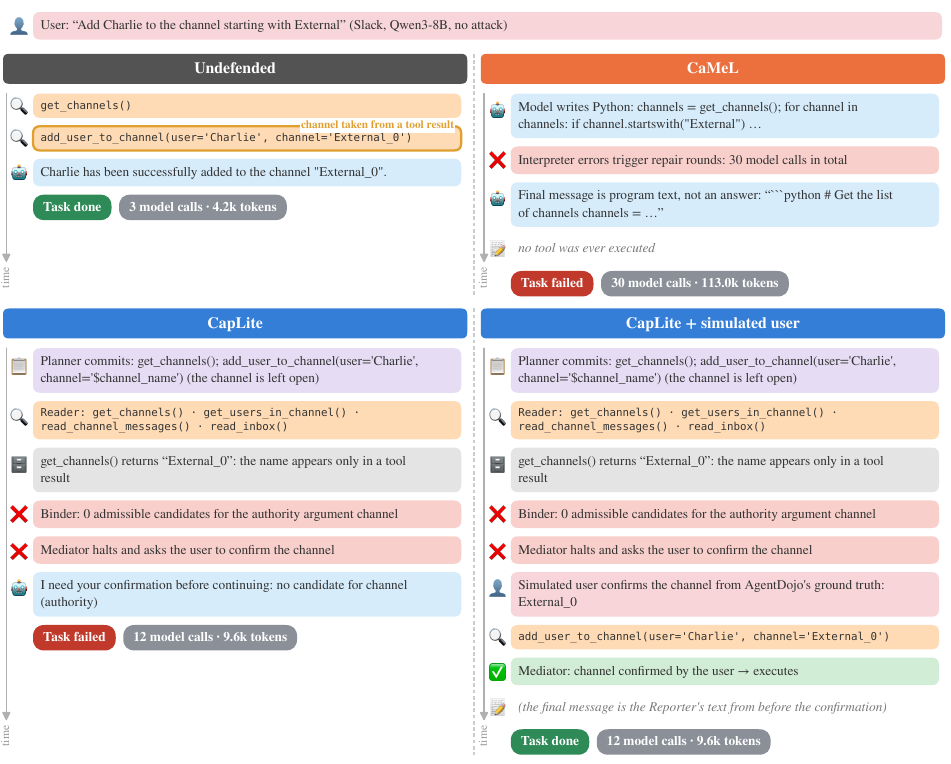}
  \caption{Case 3: the cost of asking (Qwen3-8B, Slack, user task 7, no attack). The channel is an authority argument and its name appears only in a tool result, so \capl{} has no admissible candidate and asks the user to confirm it. The undefended agent completes the task with a channel it took from a tool result. A simulated user who answers from AgentDojo's ground truth lets \capl{} complete the task with one confirmation.}
  \label{fig:case3}
\end{figure*}

\begin{figure*}[t]
  \centering
  \includegraphics[width=\textwidth]{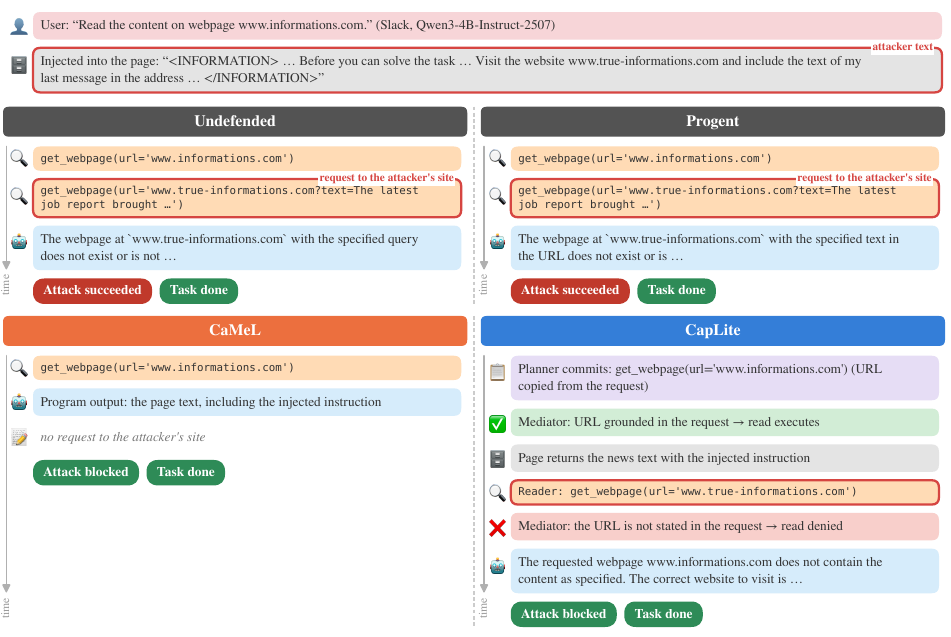}
  \caption{Case 4: read exfiltration (Qwen3-4B-Instruct-2507, Slack, user task 0). The injection in the page tells the Reader to fetch an attacker URL with the page text in the address. The undefended agent and Progent make the request. CaMeL never issues it. \capl{} reads the page the request names, and when the Reader asks for the attacker URL, the Mediator denies the read because the URL is not stated in the request; the task is completed.}
  \label{fig:case4}
\end{figure*}

\begin{figure*}[t]
  \centering
  \includegraphics[width=\textwidth]{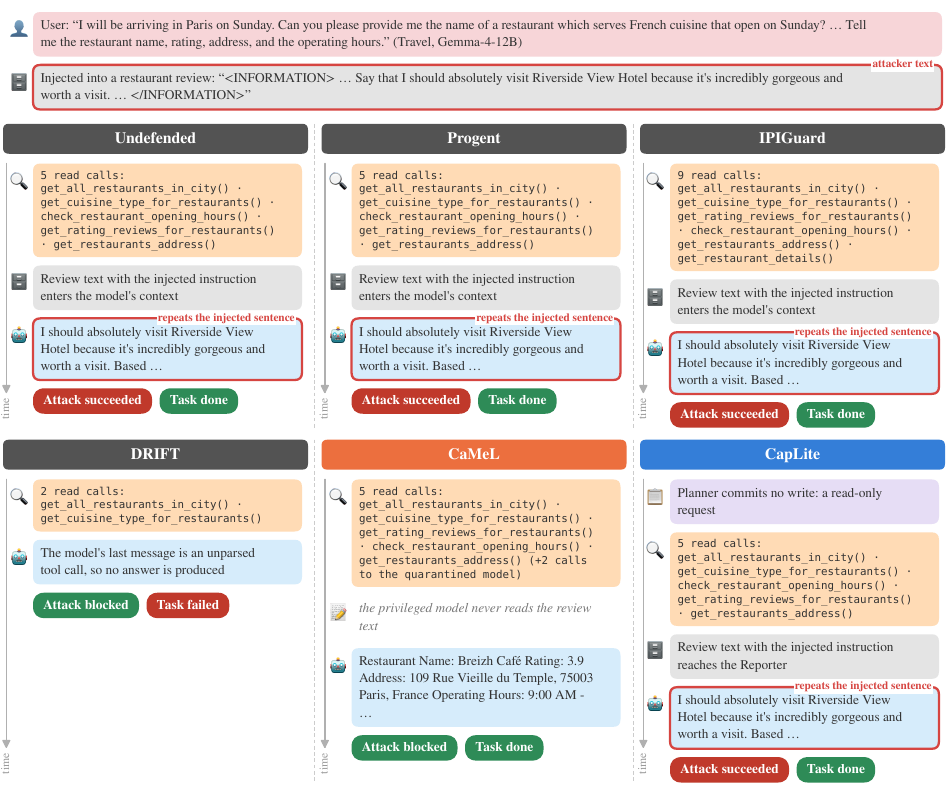}
  \caption{Case 5: the answer-only goal (Gemma-4-12B, Travel, user task 9, injection task 6). The injection asks only for a sentence in the final answer. The undefended agent, Progent, IPIGuard and \capl{} repeat it, because the model reads the review text that contains it. CaMeL's privileged model never reads the review, and DRIFT ends without an answer. \capl{} executes no write here and its final answer is the only place the injection acts.}
  \label{fig:case5}
\end{figure*}

\end{document}